\documentclass[10pt,journal,compsoc]{IEEEtran}

\usepackage[super, numbers]{natbib}
\setcitestyle{citesep={), },close={)}}
\usepackage{graphicx}
\usepackage{lipsum}

\begin{document}
%
\title{How robots in a large group make decisions as a whole? From biological inspiration to the design of distributed algorithms}

\author{Gabriele Valentini\\
School of Earth and Space Exploration\\
School of Life Sciences\\
Arizona State University, Tempe, AZ, 85827\\
gvalentini@asu.edu
}

\IEEEtitleabstractindextext{%
\begin{abstract}
Nature provides us with abundant examples of how large numbers of individuals can make decisions without the coordination of a central authority. Social insects, birds, fishes, and many other living collectives, rely on simple interaction mechanisms to do so. They individually gather information from the environment; small bits of a much larger picture that are then shared locally among the members of the collective and processed together to output a commonly agreed choice. Throughout evolution, Nature found solutions to collective decision-making problems that are intriguing to engineers for their robustness to malfunctioning or lost individuals, their flexibility in face of dynamic environments, and their ability to scale with large numbers of members. In the last decades, whereas biologists amassed large amounts of experimental evidence, engineers took inspiration from these and other examples to design distributed algorithms that, while maintaining the same properties of their natural counterparts, come with guarantees on their performance in the form of predictive mathematical models. In this paper, we review the fundamental processes that lead to a collective decision. We discuss examples of collective decisions in biological systems and show how similar processes can be engineered to design artificial ones. During this journey, we review a framework to design distributed decision-making algorithms that are modular, can be instantiated and extended in different ways, and are supported by a suit of predictive mathematical models. 
\end{abstract}

}

\maketitle

\IEEEdisplaynontitleabstractindextext
\IEEEpeerreviewmaketitle

\section{Introduction}
\label{sec:intro}

Swarm robotics focuses on the study of distributed robotics systems composed of a large number of independent and autonomous robots. Swarm engineers design these systems with three objectives in mind: \textit{i.\ scalability}, the ability of the system to keep functioning for an increasing number of components; \textit{ii.\ flexibility}, the ability of the system to adapt to unknown or changing environmental conditions; and \textit{iii.\ robustness}, the ability of the system to undergo a graceful degradation of its performance due to the loss or malfunctioning of one or more of its components~\citep{brambilla|ferrante|birattari|dorigo:2013, dorigo|birattari|brambilla:2014}. The lack of a centralized authority governing the functioning of a robot swarm is one (among others, cf.\ \citet{brambilla|ferrante|birattari|dorigo:2013}) of the fundamental pillars that allow engineers to design robotic systems that can meet these objectives. However, when no central authority is present, the swarm requires mechanisms that allow its numerous members to make decisions collectively~\citep{valentini|ferrante|dorigo:2017, valentini:2017}.

A collective decision is the result of a process distributed among a collective of agents that leads the collective to make a choice that, once made, can no longer be traced back to any of its individual agents. Collective decisions are widespread in group-living animals~\citep{conradt2003group, sumpter:2006} and are studied across different scientific disciplines such as psychology~\citep{moscovici|etal:1969,hirokawa|poole:1996}, biology~\citep{camazine|deneubourg|franks|sneyd|theraulaz|bonabeau:2001,couzin|ioannou|demirel|gross|torney|hartnett|conradt|levin|leonard:2011,conradt|list:2009}, and physics~\citep{castellano|fortunato|loreto:2009,galam:2008}, to name a few. Examples of collective decisions include the ways with which eusocial insects, such as ants and bees, explore a large portion of their environment, identify several candidate locations, and eventually choose a single, possibly optimal option where to move and create a new home~\citep{franks|pratt|mallon|britton|sumpter:2002,Visscher:2007,McGlynn:2012}; the mechanisms used by groups of vertebrates, such as bird flocks, fish schools, or primate troops, to coherently and suddenly change direction of motion in response to predators or other sources of external information~\citep{Okubo:1986,couzin|krause|franks|levin:2005,sumpter:2010,CONRADT2010675,kao|miller|torney|hartnett|couzin:2014,Strandburg-Peshkin|etal:2015}; even organisms that lack a brain, such as slime molds, are sometimes able to perform complex computation and make collective decisions~\citep{reid:2016,ray|etal:2019}. In these examples, the collective decision is represented by a consensus shared by all (or a large majority) of the members of the collective. 

Not all collective decision-making processes, however, result in a consensus among the members of the collective. Some of these processes require the collective to allocate its members to a number of different tasks (or options) in a manner that optimises performance. For example, in the realm of eusocial insects, workers take care of a number of different tasks, including foraging, brood care, and nest construction, and change their allocation as a function of colony needs, seasonality, or circadian rhythm~\citep{wiernasz2008mating,jandt|gordon2016,gordon2016,Charbonneau2015,doi:10.1098/rstb.2017.0235}. Studies of task allocation and, more generally, division of labor led to the development of different  models~\citep{beshers2001models} and ultimately inspired the design of artificial systems in different sub-fields of engineering~\citep{johnson2002emergence,gerkey|mataric:2004}.

\begin{figure}
    \centering
    \includegraphics[width=0.45\textwidth]{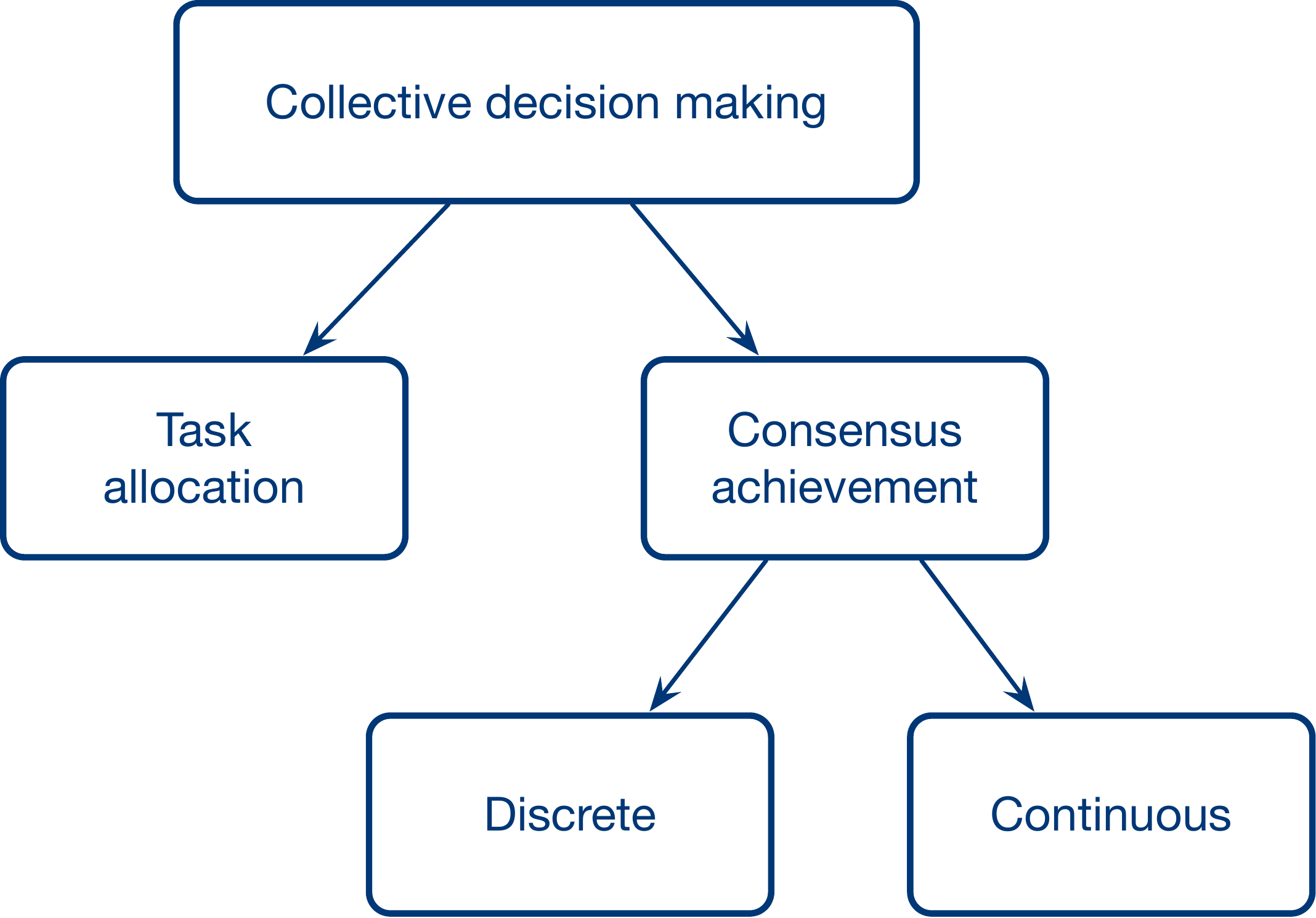}
    \caption{Hierarchical representation of different branches of collective decision making. Reproduced from~\cite{valentini|ferrante|dorigo:2017}.}
    \label{fig:cdmHierarchy}
\end{figure}

In the context of the design of robot swarms, \cite{brambilla|ferrante|birattari|dorigo:2013} distinguished these two scenarios as different categories of collective decisions (Figure~\ref{fig:cdmHierarchy}). \emph{Consensus achievement} encompasses problems that require the members of a swarm to agree on a single choice~\citep{valentini:2017}. \emph{Task allocation} includes instead problems that require the members of the swarm to allocate themselves to different tasks and, in so doing, to break down a bigger problem into a set of smaller ones that can be tackled in a manner that optimizes the performance of the swarm~\citep{gerkey|mataric:2004}. In this article, we focus on consensus achievement problems. This type of problems can be further divided into two subcategories (Figure~\ref{fig:cdmHierarchy}): problems where the swarm needs to choose one among a discrete and finite set of alternatives, i.e., to make a \textit{discrete} choice; and problems where instead the number of options is uncountable and the choice is \textit{continuous} such as the direction of motion of a swarm of agents~\citep{reynolds:1987,Olfati-Saber|Fax|Murray:2007}.

While this paper focuses on discrete consensus achievement problems in the context of robot swarms, it is worth noting that discrete and continuous consensus achievement are the subject of studies in other branches of engineering. In artificial intelligence and robotics, for example, discrete problems are considered when agents in a team need to cooperate with each other~\citep{pynadath|tambe:2002,bernstein|etal:2002} such as in the context of the RoboCup competition~\citep{kitano|etal:1997} where both centralized and decentralized solutions are studied~\citep{bowling|bowling|veloso:2004,kok|etal:2003a,kok|etal2003b}; or in the context of sensor fusion where the solution to similar problems is required to perform distributed object classification~\citep{kornienko|kornienko|constantinescu|pradier|levi:2005, kornienko|kornienko|levi:2005}. Continuous problems, which are not dealt with in this paper, are also widespread in swarm robotics both in the context of autonomous ground robots~\citep{turgut|etal:2008,nembrini|etal:2002,spears|etal:2004,ferrante|etal:2012,ferrante|etal:2014} and in that of unmanned aerial vehicles~\citep{holland|etal:2005,huert|etal:2011}. The community of control theory has also intensively studied the problem of continuous consensus  achievement~\citep{mesbahi|egerstedt:2010} for formation control~\citep{ren|beard|atkins:2005}, agreement on state variables~\citep{hatano|mesbahi:2005}, sensor fusion~\citep{ren|beard:2008}, and the selection of motion trajectories~\citep{sartoretti|hongler|deoliveira|mondada:2014}. Other engineering applications that can be regarded as continuous consensus achievement have been explored for the design of wireless sensor networks~\citep{schizas|giannakis|roumeliotis|ribeiro:2008,schizas|ribeiro|giannakis:2008} and other network scenarios~\citep{Komareji|Bouffanais:2013,shang|bouffanais:2014}. Although related to the studies presented here, the solutions devised in other branches of engineering tend to be too complex to be directly implemented on robot swarms due to high communication requirements. 

In the rest of this paper, we focus our discussion on discrete consensus achievement problems. We review the best-of-$n$ problem~\citep{valentini|ferrante|dorigo:2017}, a formalization of problems requiring discrete consensus achievement, and discuss in detail all of its possible variants. We then discuss three examples of application scenarios that can be modeled using this framework: the shortest-path problem, site-selection, and collective perception. Finally, we review a design framework that allows designers to develop collective decision-making strategies that are amenable to mathematical modeling.

\section{The Best-of-$n$ Problem}
\label{sec:best-of-n}

In the swarm robotics literature, a large number of research studies focused on a relatively few application scenarios whose accomplishment requires the swarm to solve a consensus achievement problem (e.g., the shortest-path problem in foraging scenarios, site-selection in aggregation scenarios). These application scenarios have been primarily tackled separately from each other with an application-oriented perspective that resulted in either the development of domain-specific methodologies or the design of black-box controllers~\citep{valentini|ferrante|dorigo:2017}. However, as we show in the rest of this paper, the consensus achievement problems underlying these application scenarios share a similar rationale and structure and it is possible to unify them into a unique framework: the \emph{best-of-$n$ problem}.

\begin{figure*}[t]
\centering
\includegraphics[width=0.9\textwidth]{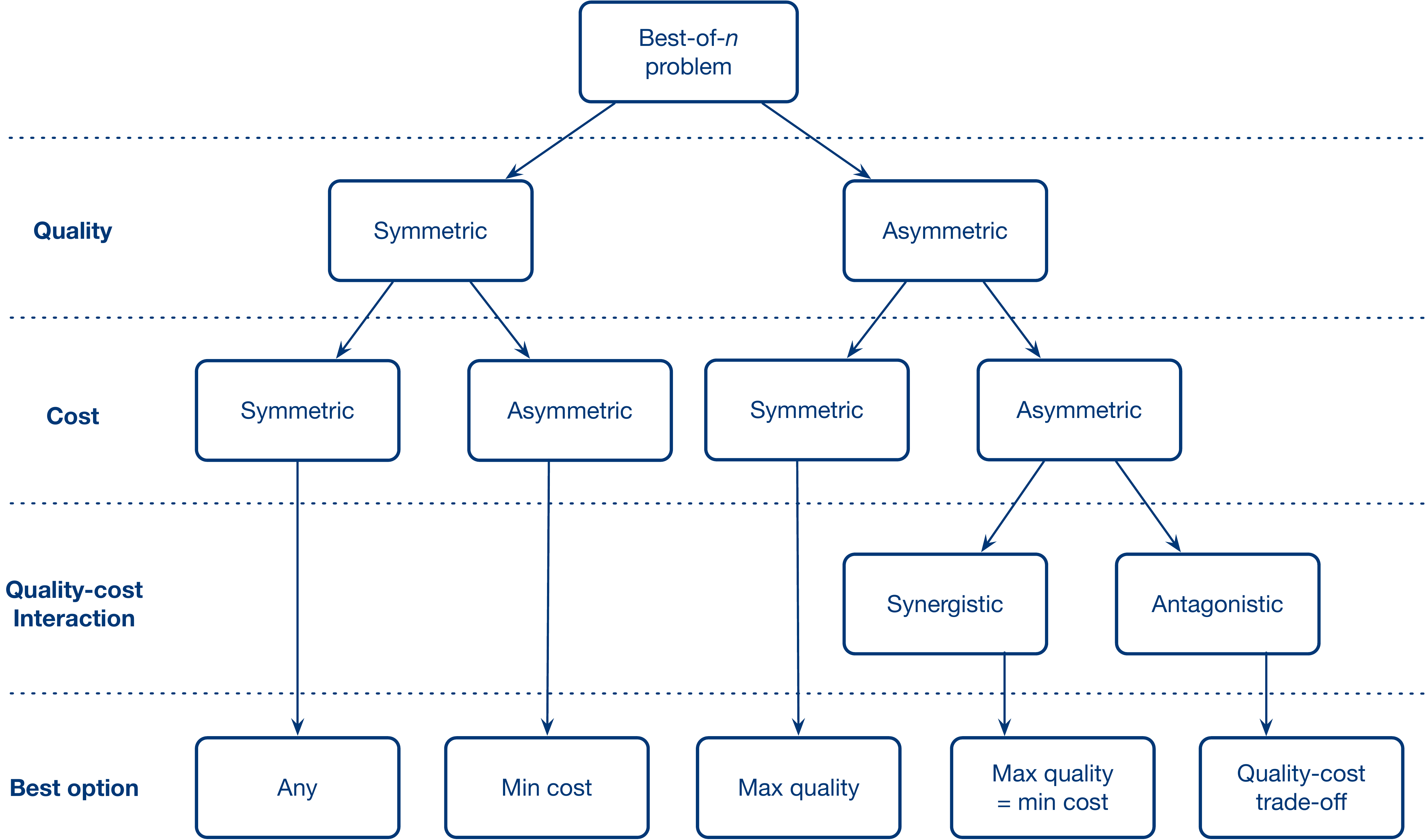}
\caption{Taxonomy of possible variants of the best-of-$n$ problem. Reproduced from~\cite{valentini|ferrante|dorigo:2017}.}
\label{fig:bestofn-classification}
\end{figure*}

The best-of-$n$ problem represents any decision-making problem in which a swarm of robots needs to make a choice for one \emph{option} among a finite set of $n$ different alternatives. Robots in the swarm have, or acquire over time, an \emph{opinion} about which of the available options represents the most beneficial choice for the swarm. This opinion usually changes over time as the robots in the swarm gather and process information about the problem at hand. The solution of the best-of-$n$ problem, i.e., a \emph{collective decision}, is represented by a large majority of the robots in the swarm sharing the same opinion. In order to maintain cohesion of the swarm, it is particularly important that the portion of the swarm agreeing on the same option is as high as possible otherwise the swarm may incur fragmentation. Consider for example an emigration problem in which options correspond to different locations in the environment. If the members of the swarm are spread over different alternatives, some of them might get lost once the emigration towards the location favored by the majority begins. In the extreme case in which all members of the swarm agree on the same option, we say that the swarm has reached a \emph{consensus} decision.

In general, different options provide the swarm with different benefits and come at different costs. Each option is therefore described by two characteristics: a \emph{quality} and a \emph{cost}. Both of them, in turn, can be a function of several attributes that depend on the specific scenario faced by the swarm~\citep{reid|garnier|beekman|latty:2015}. We borrow an example from Nature to further explain these concepts. When the reproductive season for honeybees arrives, the swarm splits and roughly half of its members emigrate to a new nest site~\citep{seeley2010honeybee}. Honeybees have a preference for sites with a certain volume, exposure to the sun, and height from the ground~\citep{camazine|visscher|finley|vetter:1999}. These attributes, possibly weighted in different ways, contribute to the perception of the site quality by individual bees. The distance of the new nest also affects the decision of the swarm as sites that are too far might be impossible to find by individual bees and sites that are too close to the original nest might result in attrition between different swarms. In this example, distance represents therefore the cost of an option while a weighted combination of the site's attributes represents its quality.

In the context of swarm robotics, the quality and the cost of each option depend on the specific target scenario and on the choices of the designer. Furthermore, depending on the ways with which quality and cost interact with each other, the definition of the optimal solution of the best-of-$n$ problem can differ (Figure~\ref{fig:bestofn-classification}).
Both quality and cost can be either \emph{symmetric}, i.e., all options are equivalent to each other, or \emph{asymmetric}, i.e., options differ in quality and/or cost. Moreover, the interaction between quality and cost for each option can be either \emph{synergistic}, i.e., producing a greater combined result, or \emph{antagonistic}, i.e., playing against each other. The combinations of these proprieties define five possible variants of the best-of-$n$ problem with different concepts of which option is the best:
\begin{enumerate}
    \item[i.] All options have the same quality and the same cost and are therefore equivalent (also know as a symmetry-breaking problem~\citep{hamann|schmickl|worn|crailsheim:2012,devries|biesmeijer:2002});
    \item[ii.] All options have the same quality but at least two of them have different cost, in which case one of the options with minimum cost is to be preferred;
    \item[iii.] All options have the same cost but at least two of them have different quality, in which case one of the options with maximum quality is to be preferred; 
    \item[iv.] Options can have different quality and different cost but the interaction between these characteristic is synergistic and the best option has maximum quality and minimum cost; 
    \item[v.] Options quality and costs differ but this time their interaction is antagonistic so that there is not a unique option to be preferred but a trade-off between different design choices.  
\end{enumerate}

\begin{figure*}[t]
\centering
\includegraphics[width=0.95\textwidth]{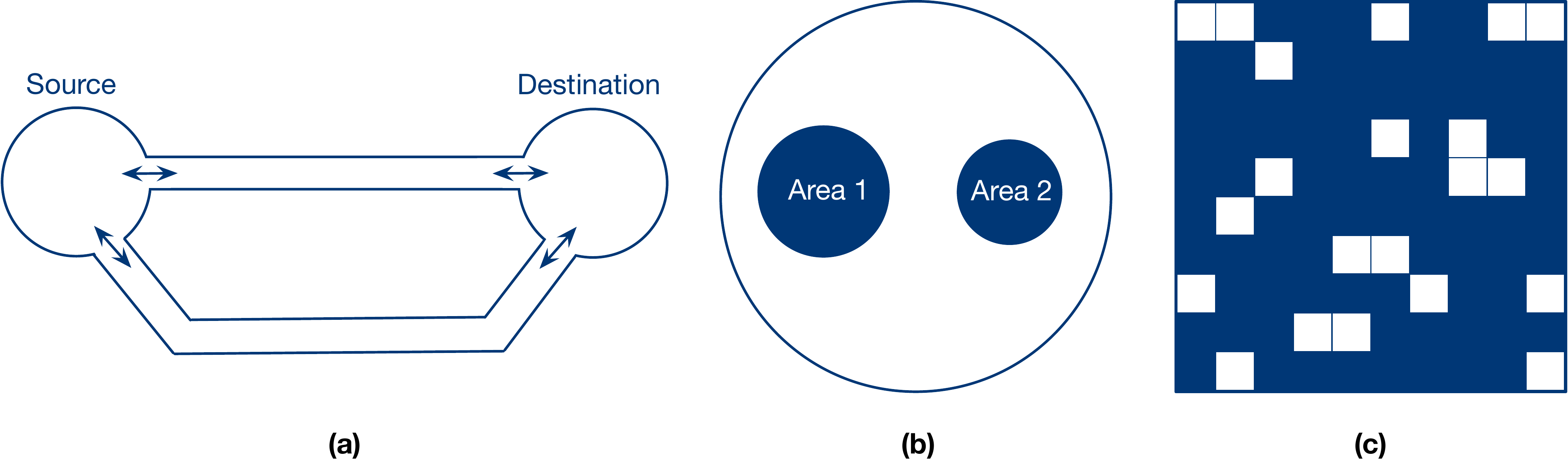}
\caption{Examples of possible scenarios that can be modeled as a best-of-$n$ problem. Panel (a) shows a shortest-path scenario with two paths of different length between a source area and a destination area. Panel (b) shows an aggregation scenario with two areas of different size. Panel (c) shows a collective perception scenario with an environment characterized by two resources represented by colors white and blue.}
\label{fig:scenarios}
\end{figure*}

Moreover, the quality and/or the cost of the available options can either be \emph{static} or vary over time and be therefore \emph{dynamic}. Static problems, which so far have been the primary subject of swarm robotics research, are usually addressed using collective decision-making strategies that result in consensus decisions~\citep{parker|zhang:2009,montesdeoca|ferrante|scheidler|pinciroli|birattari|dorigo:2011,valentini|birattari|dorigo:2013,valentini|hamann|dorigo:2014,valentini|hamann|dorigo:2015c,scheidler|brutschy|ferrante|dorigo:2015}. Dynamic problems are instead less explored in the literature of swarm robotics. In this case, to allow for the continuous exploration of new options or changes in quality and/or cost of previously discovered options, designers favor strategies that do not converge to a consensus but leave a minority of the members of the swarm unaligned with the majority~\citep{arvin|turgut|bazyari|arikan|bellotto|yue:2014, parker|zhang:2010, Prasetyo2019, CollectiveChangeDetection}.

\section{Possible application scenarios}

As introduced above, the best-of-$n$ problem is an abstraction that can be cast to a large number of different application scenarios. While we refer the reader to~\citep{valentini|ferrante|dorigo:2017} for a deeper discussion of this topic, we review here three example scenarios: shortest-path, site-selection, and collective perception. For the sake of simplicity, we present these scenarios in their binary form which can be modeled as a best-of-2 problem.

In its simplest form, the shortest-path problem is represented by a scenario in which a source area and a destination area can be reached by traversing one of two possible paths, i.e., the options of a best-of-2 problem (Figure~\ref{fig:scenarios}a). Robots in the swarm need to transport resources from the source area to the destination area. As both paths allow robots to successfully transport resources between areas, their quality is equal and symmetric. However, the efficiency with which a robot can do so might vary depending on environmental factors such as the length of a path or its roughness which affect the cost of each option. Unless options have the same cost, in which case they are all equally good, the optimal solution of the shortest-path problem is represented by the path of minimum cost. Several strategies have been developed over the years to address this problem and generally leverage the fact that the shortest path is the faster to traverse to bias the collective decision-making process~\citep{campo|etal:2010,montesdeoca|ferrante|scheidler|pinciroli|birattari|dorigo:2011,garnier|combe|jost|theraulaz:2013,valentini|birattari|dorigo:2013,scheidler|brutschy|ferrante|dorigo:2015,reina|valentini|fernandezoto|dorigo|trianni:2015,Valentini2018}. 

Another particularly popular application scenario studied in the context of the best-of-$n$ problem is site-selection~\citep{garnier|gautrais|asadpour|jost|theraulaz:2009,schmickl|thenius|moeslinger|radspieler|kernbach|szymanski|crailsheim:2009,parker|zhang:2011,valentini|hamann|dorigo:2014,valentini|hamann|dorigo:2015c,valentini|ferrante|hamann|dorigo:2016,reina|valentini|fernandezoto|dorigo|trianni:2015}. In this scenario, the swarm is located in an environment characterized by two or more sites which are of interest for some reason (e.g., the swarm may be required to choose a construction site). Sites are characterized by a quality. In the scenario depicted in Figure~\ref{fig:scenarios}b, for example, the quality of a site might be proportional to its area in which case the best site would correspond to area~1. Sites can also have an associated cost: assuming the swarm is randomly exploring the environment, area~1, being larger, is easier to discover than area~2. In this scenario, designers favor therefore strategies than can both exploit environmental biases (i.e., those associated with the cost of each option) and that can introduce a controlled bias (i.e., a mechanism that is intrinsically coded into the robot behavior) modulated by the robot's perception of the quality of a site.

Finally, an application scenario that has been recently introduced in the context of swarm robotics is that of collective perception~\citep{valentini|brambilla|hamann|dorigo:2016}. In a collective perception scenario, a swarm of robots is located in an environment characterized by different features and has to decide which feature is the most spread. In the example depicted in Figure~\ref{fig:scenarios}c, features are represented by the colors white and blue. In a real application, features may correspond to different resources available in the environment and the swarm would therefore be tasked with classifying the environment as a reservoir of the most spread resource. While the abundance of each feature represents the quality of that particular option, options could also have an associated cost (e.g., the time necessary to mine or collect a particular resource). After its introduction by~\citep{valentini|brambilla|hamann|dorigo:2016}, the collective perception scenario has been recently considered in other studies as a benchmark for collective decision making~\citep{Strobel:2018,ebert|gauci|nagpal:2018}.

\section{A design framework}

\begin{figure}[t]
    \centering
    \includegraphics[width=0.4\textwidth]{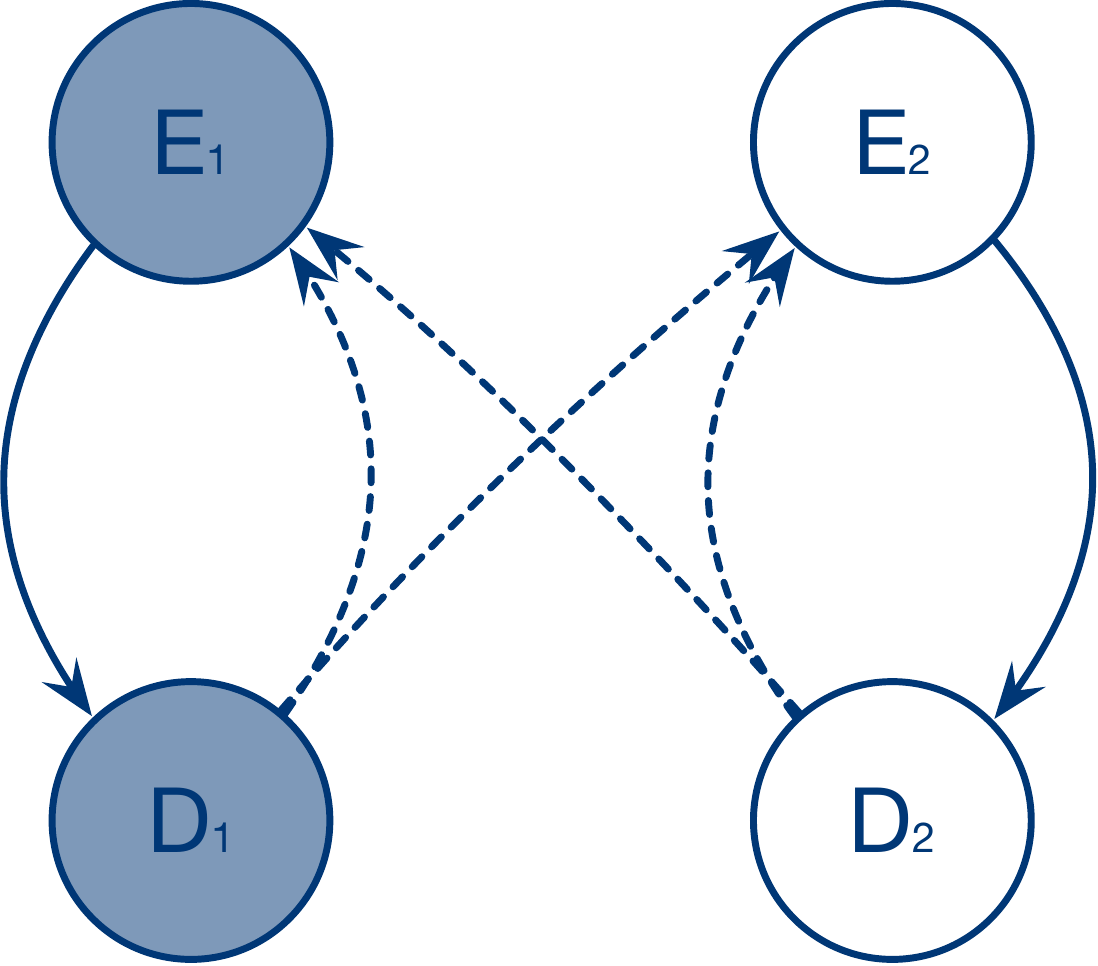}
    \caption{Finite state machine modeling a generic collective decision-making strategy for a best-of-2 problem. Colors blue and white represent option~1 and option~2; symbols \textsc{E} and \textsc{D} represent the exploration and the dissemination states; solid and dashed arrows represent deterministic and stochastic state transitions.}
    \label{fig:fsm}
\end{figure}

The best-of-$n$ problem can be address using strategies designed following different approaches (e.g., evolutionary computation). In this section, we review a behavior-based approach put forward in~\citep{valentini:2017} which aims at designing simple strategies whose dynamics are amenable to mathematical modeling. This characteristic is of particular relevance to swarm engineers as mathematical models allow designers to extensively study the performance of their strategies under different assumptions without relying on time-consuming simulations or robot experiments. Such collective decision-making strategies can be designed by combining a few key components that suffice to address all five variants of the best-of-$n$ problem. These components are: \textit{i}.\ a mechanism for \emph{exploration} of options, \textit{ii}.\ one for the \emph{dissemination} of opinions, \textit{iii}.\ and a \emph{decision rule} that allows robots to change their opinion.

Robots in the swarms need a mechanism to explore the environment. This mechanism is necessary to discover possible alternatives of the best-of-$n$ problem as well as to gather sample estimates of the associated quality. Depending on the target application scenario, exploration can be implemented as a simple random search of the environment or can combine more complex search strategies that rely on other factors, for example, that leverage the presence of a light source to drive the motion of robots using phototaxis~\citep{valentini|ferrante|hamann|dorigo:2016}. The duration of the exploration phase is generally affected by environmental factors that are beyond the reach of the designer. This point is of particular relevance as the cost of each option is generally ``paid'' in terms of the time spent exploring that option which affects the frequency with which robots can disseminate their opinion and influence the decision-making process.

Once a robot has sampled the quality of an option, it is ready to disseminate its opinion about that option within the rest of the swarm. To do so, it is important that the swarm has some sort of central location (e.g., a deployment area) where robots can exchange their opinions. Disseminating an opinion can be as simple as broadcasting it locally towards other members of the swarm. The duration of the dissemination phase is controlled by the designer and, in order to influence the collective decision towards a consensus for the best option, it needs to be proportional to the quality of the option being advertised. By doing so, options with higher quality are advertised for longer time and have higher chances to be heard by other members of the swarm.
Indeed, during the dissemination phase, a robot also listens to the opinions of other robots and keeps track of their frequency.

Before terminating a dissemination period, a robot applies a decision rule to reconsider and possibly change its own opinion. A decision rule is a function that takes as input the current robot opinion and the set of opinions perceived from neighboring robots and outputs a new opinion. To minimize the side-effects of delayed applications of a decision rule, the set of opinions of neighboring robots should be formed only by the most recently perceived opinions. Failure to do so might result in a robot applying a decision rule using outdated information possibly impacting the collective decision-making process. Popular decision rules are represented by the voter model, with which a robot adopts the opinion of a randomly chosen neighbor, and the majority rule, with which a robot adopts the opinion shared by the majority of its neighbors~\citep{valentini|ferrante|hamann|dorigo:2016}. Other decision rules are possible but an extensive discussion of this point is beyond our scope (see \citep{valentini|ferrante|dorigo:2017} for more information). 

Finally, all these mechanisms are combined using a finite state machine as illustrated in Figure~\ref{fig:fsm}. From an exploration state, \textsc{E}$_i$, in which the robot sampled the quality of option~$i$, it will deterministically transition into the dissemination state \textsc{D}$_i$ associated to the corresponding opinion. After disseminating opinion~$i$ for a time proportional to its quality, the robot applies a decision rule that is function of the most recently perceived opinions of its neighbors. The outcome of the decision rule, which depends on the composition of its neighborhood, will determine which option the robot will explore next. The execution of this finite state machine continues until the swarm reaches a collective decision. Note that repeated explorations of the the same options, even if a robot does not change opinion after applying the decision rule, are required to minimize the impact of noise during sampling of the associated quality.

\section{Conclusions}

The ability to make decisions collectively is a fundamental pillar for the development of autonomous robot swarms. In this paper, we reviewed the best-of-$n$ problem, a formal framework to model discrete consensus achievement problems for robot swarms. We showed that this framework can be used to model different application scenarios. The advantage of taking this perspective is that, once a particular strategy to address the best-of-$n$ problem is in place, it can be easily re-implemented to address a different problem scenario as showed in~\citep{valentini|brambilla|hamann|dorigo:2016}. We reviewed a design framework to devise behavior-based strategies that are amenable to mathematical modeling. This design framework is appealing to swarm engineers seeking to guarantee the performance of their distributed algorithms through formal analysis of the resulting population dynamics~\citep{valentini:2017} and/or information transfer processes~\citep{valentini|etal:2018alife}. Simple collective decision-making strategies can be later extended to incorporate more advanced mechanisms and improve performance and flexibility~\citep{talamali2019improving}.

\bibliographystyle{abbrvnat}
\bibliography{swarmrobotics}

\begin{thebibliography}{86}
\providecommand{\natexlab}[1]{#1}
\providecommand{\url}[1]{\texttt{#1}}
\expandafter\ifx\csname urlstyle\endcsname\relax
  \providecommand{\doi}[1]{doi: #1}\else
  \providecommand{\doi}{doi: \begingroup \urlstyle{rm}\Url}\fi

\bibitem[Arvin et~al.(2014)Arvin, Turgut, Bazyari, Arikan, Bellotto, and
  Yue]{arvin|turgut|bazyari|arikan|bellotto|yue:2014}
F.~Arvin, A.~E. Turgut, F.~Bazyari, K.~B. Arikan, N.~Bellotto, and S.~Yue.
\newblock Cue-based aggregation with a mobile robot swarm: a novel fuzzy-based
  method.
\newblock \emph{Adaptive Behavior}, 22\penalty0 (3):\penalty0 189--206, 2014.

\bibitem[Bernstein et~al.(2002)Bernstein, Givan, Immerman, and
  Zilberstein]{bernstein|etal:2002}
D.~S. Bernstein, R.~Givan, N.~Immerman, and S.~Zilberstein.
\newblock The complexity of decentralized control of markov decision processes.
\newblock \emph{Mathematics of Operations Research}, 27\penalty0 (4):\penalty0
  819--840, 2002.

\bibitem[Beshers and Fewell(2001)]{beshers2001models}
S.~N. Beshers and J.~H. Fewell.
\newblock Models of division of labor in social insects.
\newblock \emph{Annual review of entomology}, 46\penalty0 (1):\penalty0
  413--440, 2001.

\bibitem[Bowling et~al.(2004)Bowling, Browning, and
  Veloso]{bowling|bowling|veloso:2004}
M.~H. Bowling, B.~Browning, and M.~M. Veloso.
\newblock Plays as effective multiagent plans enabling opponent-adaptive play
  selection.
\newblock In \emph{Proceedings of the 14th International Conference on
  Automated Planning \& Scheduling}, pages 376--383. AAAI Press, 2004.

\bibitem[Brambilla et~al.(2013)Brambilla, Ferrante, Birattari, and
  Dorigo]{brambilla|ferrante|birattari|dorigo:2013}
M.~Brambilla, E.~Ferrante, M.~Birattari, and M.~Dorigo.
\newblock Swarm robotics: a review from the swarm engineering perspective.
\newblock \emph{Swarm Intelligence}, 7\penalty0 (1):\penalty0 1--41, 2013.

\bibitem[Camazine et~al.(1999)Camazine, Visscher, Finley, and
  Vetter]{camazine|visscher|finley|vetter:1999}
S.~Camazine, P.~K. Visscher, J.~Finley, and R.~S. Vetter.
\newblock House-hunting by honey bee swarms: collective decisions and
  individual behaviors.
\newblock \emph{Insectes Sociaux}, 46\penalty0 (4):\penalty0 348--360, 1999.

\bibitem[Camazine et~al.(2001)Camazine, Deneubourg, Franks, Sneyd, Theraulaz,
  and Bonabeau]{camazine|deneubourg|franks|sneyd|theraulaz|bonabeau:2001}
S.~Camazine, J.-L. Deneubourg, N.~R. Franks, J.~Sneyd, G.~Theraulaz, and
  E.~Bonabeau.
\newblock \emph{Self-Organization in Biological Systems}.
\newblock Princeton University Press, Princeton, NJ, 2001.

\bibitem[Campo et~al.(2010)Campo, Guti{\'e}rrez, Nouyan, Pinciroli, Longchamp,
  Garnier, and Dorigo]{campo|etal:2010}
A.~Campo, {\'A}.~Guti{\'e}rrez, S.~Nouyan, C.~Pinciroli, V.~Longchamp,
  S.~Garnier, and M.~Dorigo.
\newblock Artificial pheromone for path selection by a foraging swarm of
  robots.
\newblock \emph{Biological Cybernetics}, 103\penalty0 (5):\penalty0 339--352,
  2010.

\bibitem[Castellano et~al.(2009)Castellano, Fortunato, and
  Loreto]{castellano|fortunato|loreto:2009}
C.~Castellano, S.~Fortunato, and V.~Loreto.
\newblock Statistical physics of social dynamics.
\newblock \emph{Reviews of Modern Physics}, 81:\penalty0 591--646, 2009.

\bibitem[Charbonneau and Dornhaus(2015)]{Charbonneau2015}
D.~Charbonneau and A.~Dornhaus.
\newblock Workers `specialized' on inactivity: Behavioral consistency of
  inactive workers and their role in task allocation.
\newblock \emph{Behavioral Ecology and Sociobiology}, 69\penalty0 (9):\penalty0
  1459--1472, 2015.

\bibitem[Conradt and List(2009)]{conradt|list:2009}
L.~Conradt and C.~List.
\newblock Group decisions in humans and animals: {A} survey.
\newblock \emph{Philosophical Transactions of the Royal Society B: Biological
  Sciences}, 364\penalty0 (1518):\penalty0 719--742, 2009.

\bibitem[Conradt and Roper(2003)]{conradt2003group}
L.~Conradt and T.~J. Roper.
\newblock Group decision-making in animals.
\newblock \emph{Nature}, 421\penalty0 (6919):\penalty0 155, 2003.

\bibitem[Conradt and Roper(2010)]{CONRADT2010675}
L.~Conradt and T.~J. Roper.
\newblock Deciding group movements: Where and when to go.
\newblock \emph{Behavioural Processes}, 84\penalty0 (3):\penalty0 675--677,
  2010.

\bibitem[Couzin et~al.(2005)Couzin, Krause, Franks, and
  Levin]{couzin|krause|franks|levin:2005}
I.~D. Couzin, J.~Krause, N.~R. Franks, and S.~A. Levin.
\newblock Effective leadership and decision-making in animal groups on the
  move.
\newblock \emph{Nature}, 433\penalty0 (7025):\penalty0 513--516, 2005.

\bibitem[Couzin et~al.(2011)Couzin, Ioannou, Demirel, Gross, Torney, Hartnett,
  Conradt, Levin, and
  Leonard]{couzin|ioannou|demirel|gross|torney|hartnett|conradt|levin|leonard:2011}
I.~D. Couzin, C.~C. Ioannou, G.~Demirel, T.~Gross, C.~J. Torney, A.~Hartnett,
  L.~Conradt, S.~A. Levin, and N.~E. Leonard.
\newblock Uninformed individuals promote democratic consensus in animal groups.
\newblock \emph{Science}, 334\penalty0 (6062):\penalty0 1578--1580, 2011.

\bibitem[de~Vries and Biesmeijer(2002)]{devries|biesmeijer:2002}
H.~de~Vries and J.~C. Biesmeijer.
\newblock Self-organization in collective honeybee foraging: emergence of
  symmetry breaking, cross inhibition and equal harvest-rate distribution.
\newblock \emph{Behavioral Ecology and Sociobiology}, 51\penalty0 (6):\penalty0
  557--569, 2002.

\bibitem[Dorigo et~al.(2014)Dorigo, Birattari, and
  Brambilla]{dorigo|birattari|brambilla:2014}
M.~Dorigo, M.~Birattari, and M.~Brambilla.
\newblock {S}warm robotics.
\newblock \emph{Scholarpedia}, 9\penalty0 (1):\penalty0 1463, 2014.

\bibitem[Ebert et~al.(2018)Ebert, Gauci, and Nagpal]{ebert|gauci|nagpal:2018}
J.~Ebert, M.~Gauci, and R.~Nagpal.
\newblock Multi-feature collective decision making in robot swarms.
\newblock In M.~Dastani, G.~Sukthankar, E.~Andre, and S.~Koenig, editors,
  \emph{Proceedings of the 17th International Conference on Autonomous Agents
  and Multiagent Systems}, AAMAS~'18, pages 1711--1719. IFAAMAS, 2018.

\bibitem[Ferrante et~al.(2012)Ferrante, Turgut, Huepe, Stranieri, Pinciroli,
  and Dorigo]{ferrante|etal:2012}
E.~Ferrante, A.~E. Turgut, C.~Huepe, A.~Stranieri, C.~Pinciroli, and M.~Dorigo.
\newblock Self-organized flocking with a mobile robot swarm: a novel motion
  control method.
\newblock \emph{Adaptive Behavior}, 20\penalty0 (6):\penalty0 460--477, 2012.

\bibitem[Ferrante et~al.(2014)Ferrante, Turgut, Stranieri, Pinciroli,
  Birattari, and Dorigo]{ferrante|etal:2014}
E.~Ferrante, A.~E. Turgut, A.~Stranieri, C.~Pinciroli, M.~Birattari, and
  M.~Dorigo.
\newblock A self-adaptive communication strategy for flocking in stationary and
  non-stationary environments.
\newblock \emph{Natural Computing}, 13\penalty0 (2):\penalty0 225--245, 2014.

\bibitem[Franks et~al.(2002)Franks, Pratt, Mallon, Britton, and
  Sumpter]{franks|pratt|mallon|britton|sumpter:2002}
N.~R. Franks, S.~C. Pratt, E.~B. Mallon, N.~F. Britton, and D.~J.~T. Sumpter.
\newblock Information flow, opinion polling and collective intelligence in
  house-hunting social insects.
\newblock \emph{Philosophical Transactions of the Royal Society B: Biological
  Sciences}, 357\penalty0 (1427):\penalty0 1567--1583, 2002.

\bibitem[Galam(2008)]{galam:2008}
S.~Galam.
\newblock Sociophysics: A review of {G}alam models.
\newblock \emph{International Journal of Modern Physics C}, 19\penalty0
  (03):\penalty0 409--440, 2008.

\bibitem[Garnier et~al.(2009)Garnier, Gautrais, Asadpour, Jost, and
  Theraulaz]{garnier|gautrais|asadpour|jost|theraulaz:2009}
S.~Garnier, J.~Gautrais, M.~Asadpour, C.~Jost, and G.~Theraulaz.
\newblock Self-organized aggregation triggers collective decision making in a
  group of cockroach-like robots.
\newblock \emph{Adaptive Behavior}, 17\penalty0 (2):\penalty0 109--133, 2009.

\bibitem[Garnier et~al.(2013)Garnier, Combe, Jost, and
  Theraulaz]{garnier|combe|jost|theraulaz:2013}
S.~Garnier, M.~Combe, C.~Jost, and G.~Theraulaz.
\newblock Do ants need to estimate the geometrical properties of trail
  bifurcations to find an efficient route? {A} swarm robotics test bed.
\newblock \emph{PLoS Computational Biology}, 9\penalty0 (3):\penalty0 1--12,
  2013.

\bibitem[Gerkey and Matari\'c(2004)]{gerkey|mataric:2004}
B.~P. Gerkey and M.~J. Matari\'c.
\newblock A formal analysis and taxonomy of task allocation in multi-robot
  systems.
\newblock \emph{The International Journal of Robotics Research}, 23\penalty0
  (9):\penalty0 939--954, 2004.

\bibitem[Gordon(2016)]{gordon2016}
D.~M. Gordon.
\newblock From division of labor to the collective behavior of social insects.
\newblock \emph{Behavioral Ecology and Sociobiology}, 70\penalty0 (7):\penalty0
  1101--1108, 2016.

\bibitem[Hamann et~al.(2012)Hamann, Schmickl, W\"orn, and
  Crailsheim]{hamann|schmickl|worn|crailsheim:2012}
H.~Hamann, T.~Schmickl, H.~W\"orn, and K.~Crailsheim.
\newblock Analysis of emergent symmetry breaking in collective decision making.
\newblock \emph{Neural Computing and Applications}, 21\penalty0 (2):\penalty0
  207--218, 2012.

\bibitem[Hatano and Mesbahi(2005)]{hatano|mesbahi:2005}
Y.~Hatano and M.~Mesbahi.
\newblock Agreement over random networks.
\newblock \emph{IEEE Transactions on Automatic Control}, 50\penalty0
  (11):\penalty0 1867--1872, 2005.

\bibitem[Hauert et~al.(2011)Hauert, Leven, Varga, Ruini, Cangelosi, Zufferey,
  and Floreano]{huert|etal:2011}
S.~Hauert, S.~Leven, M.~Varga, F.~Ruini, A.~Cangelosi, J.~C. Zufferey, and
  D.~Floreano.
\newblock Reynolds flocking in reality with fixed-wing robots: Communication
  range vs. maximum turning rate.
\newblock In \emph{Proceedings of the 2011 IEEE/RSJ International Conference on
  Intelligent Robots and Systems, IROS 2011}, pages 5015--5020, 2011.

\bibitem[Hirokawa and Poole(1996)]{hirokawa|poole:1996}
R.~Y. Hirokawa and M.~S. Poole.
\newblock \emph{Communication and group decision making}.
\newblock Sage Publications, 1996.

\bibitem[Holland et~al.(2005)Holland, Woods, Nardi, and
  Clark]{holland|etal:2005}
O.~Holland, J.~Woods, R.~D. Nardi, and A.~Clark.
\newblock Beyond swarm intelligence: {T}he {U}ltra{S}warm.
\newblock In \emph{Proceedings of the 2005 IEEE Swarm Intelligence Symposium,
  SIS 2005}, pages 217--224, 2005.

\bibitem[Jandt and Gordon(2016)]{jandt|gordon2016}
J.~M. Jandt and D.~M. Gordon.
\newblock The behavioral ecology of variation in social insects.
\newblock \emph{Current Opinion in Insect Science}, 15:\penalty0 40--44, 2016.

\bibitem[Johnson(2002)]{johnson2002emergence}
S.~Johnson.
\newblock \emph{Emergence: The connected lives of ants, brains, cities, and
  software}.
\newblock Simon and Schuster, 2002.

\bibitem[Kao et~al.(2014)Kao, Miller, Torney, Hartnett, and
  Couzin]{kao|miller|torney|hartnett|couzin:2014}
A.~B. Kao, N.~Miller, C.~Torney, A.~Hartnett, and I.~D. Couzin.
\newblock Collective learning and optimal consensus decisions in social animal
  groups.
\newblock \emph{PLoS Computational Biology}, 10\penalty0 (8):\penalty0 1--11,
  08 2014.

\bibitem[Kitano et~al.(1997)Kitano, Asada, Kuniyoshi, Noda, Osawa, and
  Matsubara]{kitano|etal:1997}
H.~Kitano, M.~Asada, Y.~Kuniyoshi, I.~Noda, E.~Osawa, and H.~Matsubara.
\newblock {R}obo{C}up: {A} challenge problem for {AI}.
\newblock \emph{AI magazine}, 18\penalty0 (1):\penalty0 73--85, 1997.

\bibitem[Kok and Vlassis(2003)]{kok|etal2003b}
J.~R. Kok and N.~Vlassis.
\newblock Distributed decision making of robotic agents.
\newblock In \emph{Proceedings of the 8th Annual Conference of the Advanced
  School for Computing and Imaging}, pages 318--325, 2003.

\bibitem[Kok et~al.(2003)Kok, Spaan, and Vlassis]{kok|etal:2003a}
J.~R. Kok, M.~T. Spaan, and N.~Vlassis.
\newblock Multi-robot decision making using coordination graphs.
\newblock In \emph{Proceedings of the 11th International Conference on Advanced
  Robotics, ICAR}, pages 1124--1129, 2003.

\bibitem[Komareji and Bouffanais(2013)]{Komareji|Bouffanais:2013}
M.~Komareji and R.~Bouffanais.
\newblock Resilience and controllability of dynamic collective behaviors.
\newblock \emph{PLOS ONE}, 8\penalty0 (12):\penalty0 1--15, 12 2013.

\bibitem[Kornienko et~al.(2005{\natexlab{a}})Kornienko, Kornienko,
  Constantinescu, Pradier, and
  Levi]{kornienko|kornienko|constantinescu|pradier|levi:2005}
S.~Kornienko, O.~Kornienko, C.~Constantinescu, M.~Pradier, and P.~Levi.
\newblock Cognitive micro-agents: {I}ndividual and collective perception in
  microrobotic swarm.
\newblock In \emph{Proceedings of the IJCAI-05 Workshop on Agents in real-time
  and dynamic environments}, pages 33--42, 2005{\natexlab{a}}.

\bibitem[Kornienko et~al.(2005{\natexlab{b}})Kornienko, Kornienko, and
  Levi]{kornienko|kornienko|levi:2005}
S.~Kornienko, O.~Kornienko, and P.~Levi.
\newblock Minimalistic approach towards communication and perception in
  microrobotic swarms.
\newblock In \emph{2005 IEEE/RSJ International Conference on Intelligent Robots
  and Systems (IROS)}, pages 2228--2234, 2005{\natexlab{b}}.

\bibitem[Kwapich et~al.(2018)Kwapich, Valentini, and
  H\"olldobler]{doi:10.1098/rstb.2017.0235}
C.~L. Kwapich, G.~Valentini, and B.~H\"olldobler.
\newblock The non-additive effects of body size on nest architecture in a
  polymorphic ant.
\newblock \emph{Philosophical Transactions of the Royal Society B: Biological
  Sciences}, 373\penalty0 (1753):\penalty0 20170235, 2018.

\bibitem[McGlynn(2012)]{McGlynn:2012}
T.~P. McGlynn.
\newblock The ecology of nest movement in social insects.
\newblock \emph{Annual Review of Entomology}, 57\penalty0 (1):\penalty0
  291--308, 2012.
\newblock PMID: 21910641.

\bibitem[Mesbahi and Egerstedt(2010)]{mesbahi|egerstedt:2010}
M.~Mesbahi and M.~Egerstedt.
\newblock \emph{Graph theoretic methods in multiagent networks}.
\newblock Princeton Series in Applied Mathematics. Princeton University Press,
  Princeton, NJ, 2010.

\bibitem[Montes~de Oca et~al.(2011)Montes~de Oca, Ferrante, Scheidler,
  Pinciroli, Birattari, and
  Dorigo]{montesdeoca|ferrante|scheidler|pinciroli|birattari|dorigo:2011}
M.~A. Montes~de Oca, E.~Ferrante, A.~Scheidler, C.~Pinciroli, M.~Birattari, and
  M.~Dorigo.
\newblock Majority-rule opinion dynamics with differential latency: {A}
  mechanism for self-organized collective decision-making.
\newblock \emph{Swarm Intelligence}, 5:\penalty0 305--327, 2011.

\bibitem[Moscovici and Zavalloni(1969)]{moscovici|etal:1969}
S.~Moscovici and M.~Zavalloni.
\newblock The group as a polarizer of attitudes.
\newblock \emph{Journal of Personality and Social Psychology}, 12\penalty0
  (2):\penalty0 125--135, 1969.

\bibitem[Nembrini et~al.(2002)Nembrini, Winfield, and
  Melhuish]{nembrini|etal:2002}
J.~Nembrini, A.~Winfield, and C.~Melhuish.
\newblock Minimalist coherent swarming of wireless networked autonomous mobile
  robots.
\newblock In \emph{Proceedings of the Seventh International Conference on
  Simulation of Adaptive Behavior on From Animals to Animats}, ICSAB, pages
  373--382. MIT Press, 2002.

\bibitem[Okubo(1986)]{Okubo:1986}
A.~Okubo.
\newblock Dynamical aspects of animal grouping: {S}warms, schools, flocks, and
  herds.
\newblock \emph{Advances in Biophysics}, 22:\penalty0 1--94, 1986.

\bibitem[{Olfati-Saber} et~al.(2007){Olfati-Saber}, Fax, and
  Murray]{Olfati-Saber|Fax|Murray:2007}
R.~{Olfati-Saber}, J.~A. Fax, and R.~M. Murray.
\newblock Consensus and cooperation in networked multi-agent systems.
\newblock \emph{Proceedings of the IEEE}, 95\penalty0 (1):\penalty0 215--233,
  2007.

\bibitem[Parker and Zhang(2009)]{parker|zhang:2009}
C.~A.~C. Parker and H.~Zhang.
\newblock Cooperative decision-making in decentralized multiple-robot systems:
  The best-of-n problem.
\newblock \emph{IEEE/ASME Transactions on Mechatronics}, 14\penalty0
  (2):\penalty0 240--251, 2009.

\bibitem[Parker and Zhang(2010)]{parker|zhang:2010}
C.~A.~C. Parker and H.~Zhang.
\newblock Collective unary decision-making by decentralized multiple-robot
  systems applied to the task-sequencing problem.
\newblock \emph{Swarm Intelligence}, 4:\penalty0 199--220, 2010.

\bibitem[Parker and Zhang(2011)]{parker|zhang:2011}
C.~A.~C. Parker and H.~Zhang.
\newblock Biologically inspired collective comparisons by robotic swarms.
\newblock \emph{The International Journal of Robotics Research}, 30\penalty0
  (5):\penalty0 524--535, 2011.

\bibitem[Prasetyo et~al.(2019)Prasetyo, De~Masi, and Ferrante]{Prasetyo2019}
J.~Prasetyo, G.~De~Masi, and E.~Ferrante.
\newblock Collective decision making in dynamic environments.
\newblock \emph{Swarm Intelligence}, 2019.

\bibitem[Pynadath and Tambe(2002)]{pynadath|tambe:2002}
D.~V. Pynadath and M.~Tambe.
\newblock The communicative multiagent team decision problem: {A}nalyzing
  teamwork theories and models.
\newblock \emph{Journal of Artificial Intelligence Research}, 16\penalty0
  (1):\penalty0 389--423, 2002.

\bibitem[Ray et~al.(2019)Ray, Valentini, Shah, Haque, Reid, Weber, and
  Garnier]{ray|etal:2019}
S.~K. Ray, G.~Valentini, P.~Shah, A.~Haque, C.~R. Reid, G.~F. Weber, and
  S.~Garnier.
\newblock Information transfer during food choice in the slime mold physarum
  polycephalum.
\newblock \emph{Frontiers in Ecology and Evolution}, 7:\penalty0 67, 2019.

\bibitem[Reid and Latty(2016)]{reid:2016}
C.~R. Reid and T.~Latty.
\newblock Collective behaviour and swarm intelligence in slime moulds.
\newblock \emph{FEMS Microbiology Reviews}, 40\penalty0 (6):\penalty0 798--806,
  08 2016.

\bibitem[Reid et~al.(2015)Reid, Garnier, Beekman, and
  Latty]{reid|garnier|beekman|latty:2015}
C.~R. Reid, S.~Garnier, M.~Beekman, and T.~Latty.
\newblock Information integration and multiattribute decision making in
  non-neuronal organisms.
\newblock \emph{Animal Behaviour}, 100\penalty0 (0):\penalty0 44--50, 2015.

\bibitem[Reina et~al.(2015)Reina, Valentini, Fern\'andez-Oto, Dorigo, and
  Trianni]{reina|valentini|fernandezoto|dorigo|trianni:2015}
A.~Reina, G.~Valentini, C.~Fern\'andez-Oto, M.~Dorigo, and V.~Trianni.
\newblock A design pattern for decentralised decision making.
\newblock \emph{PLoS ONE}, 10\penalty0 (10):\penalty0 e0140950, 2015.

\bibitem[Ren and Beard(2008)]{ren|beard:2008}
W.~Ren and R.~W. Beard.
\newblock \emph{Distributed consensus in multi-vehicle cooperative control:
  {T}heory and applications}.
\newblock Communications and control engineering. Springer, London, UK, 2008.

\bibitem[Ren et~al.(2005)Ren, Beard, and Atkins]{ren|beard|atkins:2005}
W.~Ren, R.~W. Beard, and E.~M. Atkins.
\newblock A survey of consensus problems in multi-agent coordination.
\newblock In \emph{Proceedings of the 2005 American Control Conference},
  volume~3, pages 1859--1864. IEEE Press, 2005.

\bibitem[Reynolds(1987)]{reynolds:1987}
C.~W. Reynolds.
\newblock Flocks, herds and schools: {A} distributed behavioral model.
\newblock \emph{ACM SIGGRAPH Computer Graphics}, 21\penalty0 (4):\penalty0
  25--34, 1987.

\bibitem[Sartoretti et~al.(2014)Sartoretti, Hongler, {d}e Oliveira, and
  Mondada]{sartoretti|hongler|deoliveira|mondada:2014}
G.~Sartoretti, M.-O. Hongler, M.~{d}e Oliveira, and F.~Mondada.
\newblock Decentralized self-selection of swarm trajectories: from dynamical
  systems theory to robotic implementation.
\newblock \emph{Swarm Intelligence}, 8\penalty0 (4):\penalty0 329--351, 2014.

\bibitem[Scheidler et~al.(2016)Scheidler, Brutschy, Ferrante, and
  Dorigo]{scheidler|brutschy|ferrante|dorigo:2015}
A.~Scheidler, A.~Brutschy, E.~Ferrante, and M.~Dorigo.
\newblock The $k$-unanimity rule for self-organized decision-making in swarms
  of robots.
\newblock \emph{IEEE Transactions on Cybernetics}, 46\penalty0 (5):\penalty0
  1175--1188, 2016.

\bibitem[Schizas et~al.(2008{\natexlab{a}})Schizas, Giannakis, Roumeliotis, and
  Ribeiro]{schizas|giannakis|roumeliotis|ribeiro:2008}
I.~Schizas, G.~Giannakis, S.~Roumeliotis, and A.~Ribeiro.
\newblock Consensus in ad hoc {WSN}s with noisy links --- {P}art {II}:
  {D}istributed estimation and smoothing of random signals.
\newblock \emph{IEEE Transactions on Signal Processing}, 56\penalty0
  (4):\penalty0 1650--1666, 2008{\natexlab{a}}.

\bibitem[Schizas et~al.(2008{\natexlab{b}})Schizas, Ribeiro, and
  Giannakis]{schizas|ribeiro|giannakis:2008}
I.~Schizas, A.~Ribeiro, and G.~Giannakis.
\newblock Consensus in ad hoc {WSN}s with noisy links --- {P}art {I}:
  {D}istributed estimation of deterministic signals.
\newblock \emph{IEEE Transactions on Signal Processing}, 56\penalty0
  (1):\penalty0 350--364, 2008{\natexlab{b}}.

\bibitem[Schmickl et~al.(2009)Schmickl, Thenius, Moeslinger, Radspieler,
  Kernbach, Szymanski, and
  Crailsheim]{schmickl|thenius|moeslinger|radspieler|kernbach|szymanski|crailsheim:2009}
T.~Schmickl, R.~Thenius, C.~Moeslinger, G.~Radspieler, S.~Kernbach,
  M.~Szymanski, and K.~Crailsheim.
\newblock Get in touch: cooperative decision making based on robot-to-robot
  collisions.
\newblock \emph{Autonomous Agents and Multi-Agent Systems}, 18\penalty0
  (1):\penalty0 133--155, 2009.

\bibitem[Seeley(2010)]{seeley2010honeybee}
T.~D. Seeley.
\newblock \emph{Honeybee democracy}.
\newblock Princeton University Press, 2010.

\bibitem[Shang and Bouffanais(2014)]{shang|bouffanais:2014}
Y.~Shang and R.~Bouffanais.
\newblock Influence of the number of topologically interacting neighbors on
  swarm dynamics.
\newblock \emph{Scientific Reports}, 4:\penalty0 4184 EP, 2014.

\bibitem[Spears et~al.(2004)Spears, Spears, Hamann, and Heil]{spears|etal:2004}
W.~M. Spears, D.~F. Spears, J.~C. Hamann, and R.~Heil.
\newblock Distributed, physics-based control of swarms of vehicles.
\newblock \emph{Autonomous Robots}, 17\penalty0 (2):\penalty0 137--162, 2004.

\bibitem[{Strandburg-Peshkin} et~al.(2015){Strandburg-Peshkin}, Farine, Couzin,
  and Crofoot]{Strandburg-Peshkin|etal:2015}
A.~{Strandburg-Peshkin}, D.~R. Farine, I.~D. Couzin, and M.~C. Crofoot.
\newblock Shared decision-making drives collective movement in wild baboons.
\newblock \emph{Science}, 348\penalty0 (6241):\penalty0 1358--1361, 2015.

\bibitem[Strobel et~al.(2018)Strobel, Castell\'{o}~Ferrer, and
  Dorigo]{Strobel:2018}
V.~Strobel, E.~Castell\'{o}~Ferrer, and M.~Dorigo.
\newblock Managing byzantine robots via blockchain technology in a swarm
  robotics collective decision making scenario.
\newblock In \emph{Proceedings of the 17th International Conference on
  Autonomous Agents and MultiAgent Systems}, AAMAS '18, pages 541--549, 2018.

\bibitem[Sumpter(2006)]{sumpter:2006}
D.~J.~T. Sumpter.
\newblock The principles of collective animal behaviour.
\newblock \emph{Philosophical Transactions of the Royal Society B: Biological
  Sciences}, 361\penalty0 (1465):\penalty0 5--22, 2006.

\bibitem[Sumpter(2010)]{sumpter:2010}
D.~J.~T. Sumpter.
\newblock \emph{Collective animal behavior}.
\newblock Princeton University Press, Princeton, NJ, 2010.

\bibitem[Talamali et~al.(2019)Talamali, Marshall, Bose, and
  Reina]{talamali2019improving}
M.~S. Talamali, J.~A.~R. Marshall, T.~Bose, and A.~Reina.
\newblock Improving collective decision accuracy via time-varying
  cross-inhibition.
\newblock In \emph{2019 IEEE International Conference on Robotics and
  Automation}. IEEE, 2019.

\bibitem[Turgut et~al.(2008)Turgut, {\c{C}}elikkanat, G{\"o}k{\c{c}}e, and
  {\c{S}}ahin]{turgut|etal:2008}
A.~E. Turgut, H.~{\c{C}}elikkanat, F.~G{\"o}k{\c{c}}e, and E.~{\c{S}}ahin.
\newblock Self-organized flocking in mobile robot swarms.
\newblock \emph{Swarm Intelligence}, 2\penalty0 (2):\penalty0 97--120, 2008.

\bibitem[Valentini(2017)]{valentini:2017}
G.~Valentini.
\newblock \emph{Achieving Consensus in Robot Swarms: Design and Analysis of
  Strategies for the best-of-n Problem}.
\newblock Springer International Publishing, Cham, Switzerland, 2017.

\bibitem[Valentini et~al.(2013)Valentini, Birattari, and
  Dorigo]{valentini|birattari|dorigo:2013}
G.~Valentini, M.~Birattari, and M.~Dorigo.
\newblock Majority rule with differential latency: An absorbing {M}arkov chain
  to model consensus.
\newblock In T.~Gilbert, M.~Kirkilionis, and G.~Nicolis, editors,
  \emph{Proceedings of the European Conference on Complex Systems 2012},
  Springer Proceedings in Complexity, pages 651--658. Springer, 2013.

\bibitem[Valentini et~al.(2014)Valentini, Hamann, and
  Dorigo]{valentini|hamann|dorigo:2014}
G.~Valentini, H.~Hamann, and M.~Dorigo.
\newblock Self-organized collective decision making: The weighted voter model.
\newblock In A.~Lomuscio, P.~Scerri, A.~Bazzan, and M.~Huhns, editors,
  \emph{Proceedings of the 13th International Conference on Autonomous Agents
  and Multiagent Systems}, AAMAS~'14, pages 45--52. IFAAMAS, 2014.

\bibitem[Valentini et~al.(2015)Valentini, Hamann, and
  Dorigo]{valentini|hamann|dorigo:2015c}
G.~Valentini, H.~Hamann, and M.~Dorigo.
\newblock Efficient decision-making in a self-organizing robot swarm: On the
  speed versus accuracy trade-off.
\newblock In R.~Bordini, E.~Elkind, G.~Weiss, and P.~Yolum, editors,
  \emph{Proceedings of the 14th International Conference on Autonomous Agents
  and Multiagent Systems}, AAMAS '15, pages 1305--1314. IFAAMAS, 2015.

\bibitem[Valentini et~al.(2016{\natexlab{a}})Valentini, Brambilla, Hamann, and
  Dorigo]{valentini|brambilla|hamann|dorigo:2016}
G.~Valentini, D.~Brambilla, H.~Hamann, and M.~Dorigo.
\newblock Collective perception of environmental features in a robot swarm.
\newblock In M.~Dorigo, M.~Birattari, X.~Li, M.~{L\'opez-Ib\'a\~nez},
  K.~Ohkura, C.~Pinciroli, and T.~St\"utzle, editors, \emph{Swarm
  Intelligence}, volume 9882 of \emph{LNCS}, pages 65--76. Springer,
  2016{\natexlab{a}}.

\bibitem[Valentini et~al.(2016{\natexlab{b}})Valentini, Ferrante, Hamann, and
  Dorigo]{valentini|ferrante|hamann|dorigo:2016}
G.~Valentini, E.~Ferrante, H.~Hamann, and M.~Dorigo.
\newblock Collective decision with 100 {K}ilobots: {S}peed versus accuracy in
  binary discrimination problems.
\newblock \emph{Autonomous Agents and Multi-Agent Systems}, 30\penalty0
  (3):\penalty0 553--580, 2016{\natexlab{b}}.

\bibitem[Valentini et~al.(2017)Valentini, Ferrante, and
  Dorigo]{valentini|ferrante|dorigo:2017}
G.~Valentini, E.~Ferrante, and M.~Dorigo.
\newblock The best-of-\emph{n} problem in robot swarms: {F}ormalization, state
  of the art, and novel perspectives.
\newblock \emph{Frontiers in Robotics and AI}, 4:\penalty0 9, 2017.

\bibitem[Valentini et~al.(2018{\natexlab{a}})Valentini, Antoun, Trabattoni,
  Wiandt, Tamura, Hocquard, Trianni, and Dorigo]{Valentini2018}
G.~Valentini, A.~Antoun, M.~Trabattoni, B.~Wiandt, Y.~Tamura, E.~Hocquard,
  V.~Trianni, and M.~Dorigo.
\newblock Kilogrid: a novel experimental environment for the kilobot robot.
\newblock \emph{Swarm Intelligence}, 12\penalty0 (3):\penalty0 245--266, Sep
  2018{\natexlab{a}}.

\bibitem[Valentini et~al.(2018{\natexlab{b}})Valentini, Moore, Hanson, Pavlic,
  Pratt, and Walker]{valentini|etal:2018alife}
G.~Valentini, D.~G. Moore, J.~R. Hanson, T.~P. Pavlic, S.~C. Pratt, and S.~I.
  Walker.
\newblock Transfer of information in collective decisions by artificial agents.
\newblock \emph{The 2019 Conference on Artificial Life}, \penalty0
  (30):\penalty0 641--648, 2018{\natexlab{b}}.

\bibitem[Visscher(2007)]{Visscher:2007}
P.~K. Visscher.
\newblock Group decision making in nest-site selection among social insects.
\newblock \emph{Annual Review of Entomology}, 52\penalty0 (1):\penalty0
  255--275, 2007.

\bibitem[Wahby et~al.(2019)Wahby, Petzold, Eschke, Schmickl, and
  Hamann]{CollectiveChangeDetection}
M.~Wahby, J.~Petzold, C.~Eschke, T.~Schmickl, and H.~Hamann.
\newblock Collective change detection: Adaptivity to dynamic swarm densities
  and light conditions in robot swarms.
\newblock \emph{The 2019 Conference on Artificial Life}, \penalty0
  (31):\penalty0 642--649, 2019.

\bibitem[Wiernasz et~al.(2008)Wiernasz, Hines, Parker, and
  Cole]{wiernasz2008mating}
D.~C. Wiernasz, J.~Hines, D.~G. Parker, and B.~J. Cole.
\newblock Mating for variety increases foraging activity in the harvester ant,
  pogonomyrmex occidentalis.
\newblock \emph{Molecular Ecology}, 17\penalty0 (4):\penalty0 1137--1144, 2008.

\end{thebibliography}

\begin{IEEEbiography}[{\includegraphics[width=1in,height=1.25in,clip,keepaspectratio]{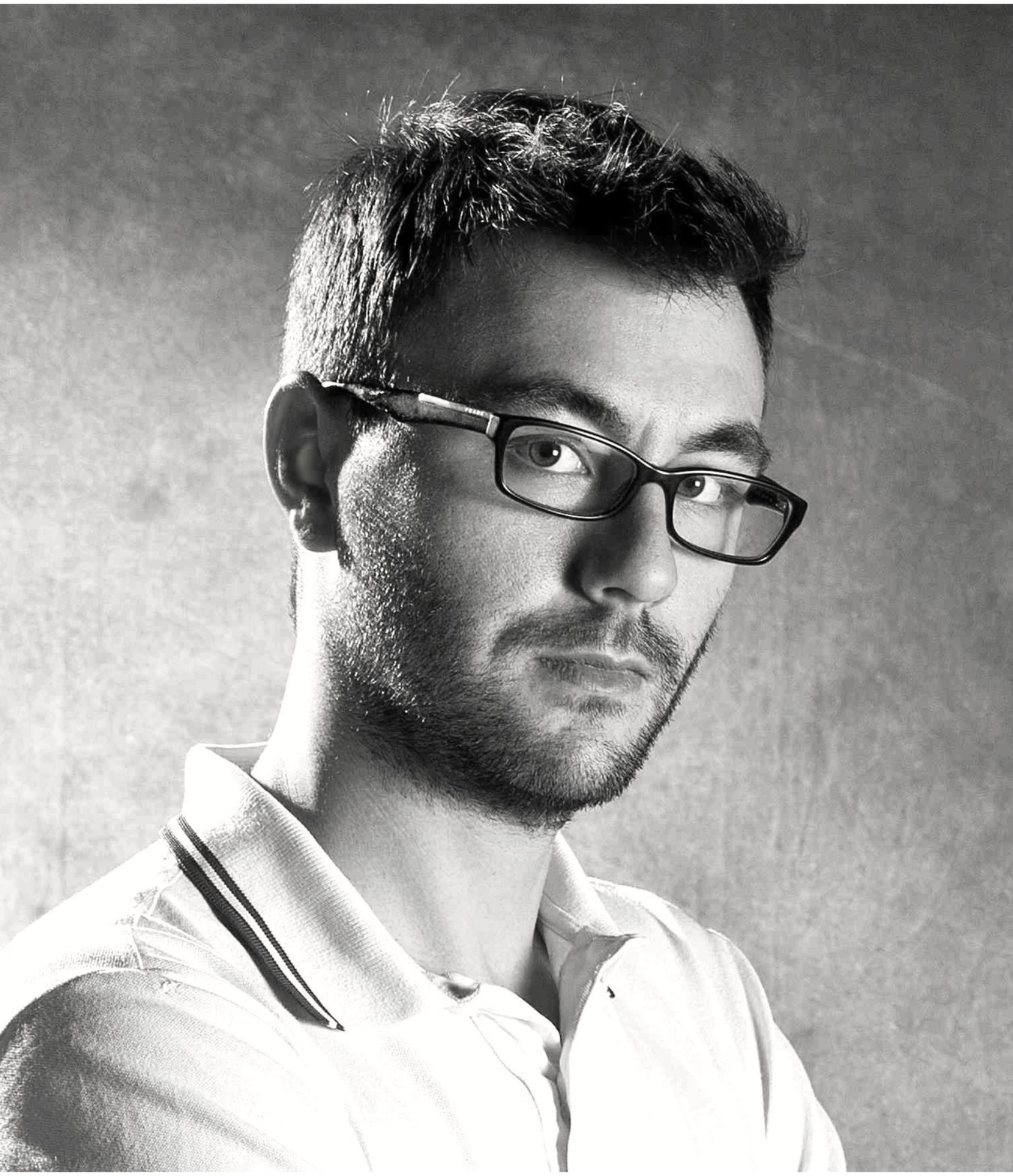}}]%
{Dr.\ Gabriele Valentini} is a Postdoctoral Research Associate at Arizona State University, Tempe, Arizona, USA. He holds a B.Sc.\ and a M.Sc.\ in Engineering of Computing Systems from Politecnico di Milano, Milan, Italy, and a Ph.D.\ in Sciences of Engineering and Technologies from Universit\'e Libre de Bruxelles, Brussels, Belgium. His doctoral dissertation has been awarded the 2016 distinguished dissertation award from the European association for Artificial Intelligence (EurAI) and the 2015 Best Student Video Award from the Association for the Advancement of Artificial Intelligence (AAAI). Dr.\ Valentini focuses on understanding and designing mechanisms for collective decision making and collective behavior both in artificial and living collectives.
\end{IEEEbiography}
\end{document}